\documentclass[]{heygen_research}

\usepackage[toc,page,header]{appendix}
\usepackage{natbib}

\usepackage{amsmath}
\usepackage{amssymb}
\usepackage{dsfont}

\usepackage{cleveref}

\usepackage{subcaption}

\usepackage{xargs}
\usepackage{todonotes}

\usepackage{multirow}

\usepackage{algorithm}
\usepackage{algorithmic}

\usepackage{array}
\usepackage{tabularx}

\usepackage{tcolorbox}

\usepackage{adjustbox}

\usepackage{bm}
\usepackage{mathrsfs}

\usepackage{adjustbox}
\usepackage{multicol}
\usepackage{changepage}
\usepackage{graphicx}
\usepackage{array}
\usepackage{hyperref}

\usepackage{minitoc}

\title{Generate Your Talking Avatar from Video Reference}

\author[1,3]{Zujin Guo}
\author[3]{Zhenhui Ye}
\author[3]{Yi Ren}
\author[3]{Yuanming Li}
\author[2,3]{Ce Chen}
\author[3]{Zhibin Hong}
\author[1]{Chen Change Loy}

\affiliation[1]{Nanyang Technological University}
\affiliation[2]{University of Melbourne}
\affiliation[3]{HeyGen Research}

\abstract{
Existing talking avatar methods typically adopt an image-to-video pipeline conditioned on a static reference image within the same scene as the target generation.
This restricted, single-view perspective lacks sufficient temporal and expression cues, limiting the ability to synthesize high-fidelity talking avatars in customized backgrounds.
To this end, we introduce Talking Avatar generation from Video Reference (TAVR), a novel framework that shifts the paradigm by leveraging cross-scene video inputs. 
To effectively process these extended temporal contexts and bridge cross-scene domain gaps, TAVR integrates a token selection module alongside a comprehensive three-stage training scheme. Specifically, same-scene video pretraining establishes foundational appearance copying, which is subsequently expanded by cross-scene reference fine-tuning for robust cross-scene adaptation. Finally, task-specific reinforcement learning aligns the generated outputs with identity-based rewards to maximize identity similarity.
To systematically evaluate cross-scene robustness, we construct a new benchmark comprising 158 carefully curated cross-scene video pairs. Extensive experiments show that TAVR benefits from flexible inference-time video referencing and consistently surpasses existing baselines both quantitatively and qualitatively. 
}
\checkdata[Project Page]{\url{https://gseancdat.github.io/projects/TAVR}}

\begin{document}

\maketitle
\let\thefootnote\relax\footnotetext{This work has been deployed to production. For more related research, please visit \href{https://www.heygen.com/research}{HeyGen Research} and \href{https://www.heygen.com/research/avatar-v-model}{HeyGen Avatar-V}.}


\section{Introduction}
Generating high-fidelity, realistic talking avatars supports a wide range of applications from digital humans~\citep{gan2025omniavatar,tu2025stableavatar} to virtual production~\citep{meng2025echomimicv3,chen2025humo}. 
Unlike general video generation, talking avatar synthesis requires a delicate balance of generating temporally coherent, speech-driven facial dynamics, such as precise lip synchronization and natural head movements, while strictly preserving the avatar's identity across varied environments.

\begin{figure*}[!t]
  \centering
  \includegraphics[width=\textwidth]{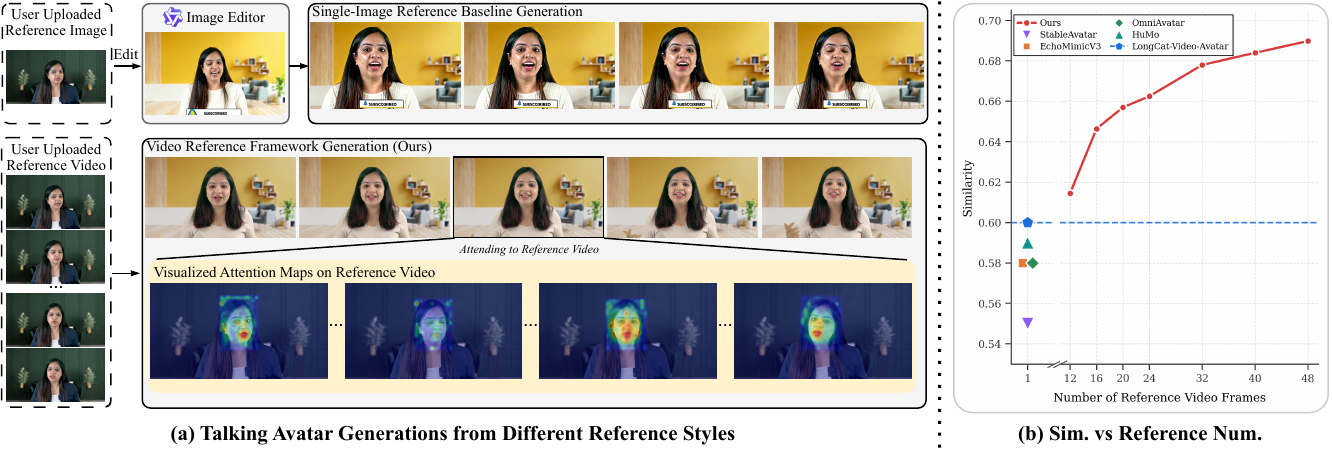}
  \caption{\textbf{Talking avatar generation from video reference.} \textbf{(a)} Visual comparisons demonstrate that our video reference framework yields significantly better identity preservation compared to the single-image baseline in cross-scene generation. To illustrate this mechanism, we visualize the attention distribution of a specific generated frame across all reference frames, obtained by extracting the attention weights directly from the model's attention layers. The heatmaps reveal that our method selectively aggregates salient identity cues (e.g., lip shapes and facial silhouettes) from highly correlated frames, while naturally suppressing frames with mismatched poses and expressions. This targeted feature aggregation explicitly demonstrates the effectiveness of the video referencing approach. \textbf{(b)} The plot shows that our identity similarity increases with the number of reference frames, confirming that longer reference input is advantageous. Furthermore, our approach consistently outperforms existing single-image baselines, achieving substantially higher similarity scores.}
  \label{fig:teaser}
\end{figure*}

Despite recent progress in diffusion models~\citep {blattmann2023svd,wan2025wan,cai2025omnivcus}, maintaining this balance remains highly challenging. Speech introduces substantial variations in facial expressions, head poses, and appearance.
Recent frameworks~\citep{cui2025hallo3,meituanlongcatteam2025longcatvideoavatartechnicalreport} typically attempt to anchor the identity of the avatar using a single reference image.
However, a single static frame lacks sufficient dynamic identity cues to maintain consistency under pose and expression changes, often resulting in immediate identity loss or identity drift during generation (Figure~\ref{fig:teaser}(a)).
Furthermore, these single-image approaches~\citep{tu2025stableavatar,gan2025omniavatar,meng2025echomimicv3} frequently struggle with cross-scene generation, \textit{i.e.}, synthesizing the avatar in a customized background different from the reference. Current cross-scene generation commonly relies on a cumbersome two-stage pipeline: image-to-image scene editing~\citep{wu2025qwen-image} followed by image-to-video synthesis. This introduces considerable computational overhead and accumulates errors, \textit{e.g.,} identity loss or excessive smoothing during the editing phase, severely limiting real-world applications. 

In this study, we shift from static image references to cross-scene video references. Designed to effectively exploit temporally-enriched visual cues, Talking Avatar generation from Video Reference (TAVR) addresses the core challenges of talking avatar synthesis within a unified framework. By leveraging multi-frame video references that capture diverse facial angles and contextual variations, our method achieves \textit{robust identity preservation} (Figure~\ref{fig:teaser}(a)), ensuring the avatar's appearance remains consistent across challenging poses and expressions, while simultaneously maintaining precise lip synchronization and synthesizing natural movements. Furthermore, TAVR allows for seamless avatar generation within \textit{customized backgrounds}, bypassing the computational redundancy inherent to the two-stage scene-editing pipelines.

Effectively leveraging cross-scene video references poses unique challenges. 
First, video inputs contain a substantially longer temporal context, which not only incurs significant computational overhead but also makes it nontrivial for the model to identify and prioritize the most informative identity features from vast amounts of visual data.
Second, cross-scene references create a pronounced domain gap with respect to the target generation, as appearance, pose, and illumination may vary significantly. 
To address these dual challenges, TAVR adopts a token selection module alongside a comprehensive three-stage training scheme. Specifically, the token selection module efficiently filters the extended temporal context to isolate the most salient identity cues. 
Concurrently, our training scheme bridges the cross-scene domain gap. The process begins with the same-scene video pretraining to establish a strong foundation for facial appearance copying. Building on this capability, we conduct cross-scene reference fine-tuning to equip the model with robust domain adaptation. In the final stage, task-specific reinforcement learning is applied to maximize the identity similarity of the generated outputs with preference rewards. Owing to this architecture and adaptive training strategy, TAVR natively supports flexible referencing at inference time, allowing the model to dynamically leverage varying lengths of cross-scene identity evidence for optimal generation (Figure~\ref{fig:teaser}(b)).

To systematically evaluate talking avatar generation under cross-scene references, we construct a new benchmark comprising 158 cross-scene video pairs curated from the TalkVid dataset~\citep{chen2025talkvid} through a dedicated data processing pipeline. Each pair features a consistent speaking identity across environments, with noticeable variations in appearance. We anticipate this benchmark to serve as a valuable resource for advancing research on robust talking avatar generation.

Our contributions are threefold: 
(i) We present TAVR, an effective talking avatar generation framework that leverages cross-scene video references to achieve robust identity preservation, customized backgrounds with precise lip synchronization and natural movements.
(ii) We propose a three-stage training paradigm that enables the model to learn informative identity features from long video contexts while adapting to cross-scene domain shifts.
(iii) We introduce a new benchmark tailored to cross-scene talking avatar generation and demonstrate through extensive experiments that TAVR consistently outperforms existing baselines on both quantitative metrics and visual quality.

\section{Related Work}
\noindent\textbf{Reference-based Video Generation.} 
Modern advances in video generation have leveraged visual reference information to improve controllability and fidelity~\citep{blattmann2023svd,wan2025wan,cheng2025wananimate,cai2025omnivcus,li2025bindweave}. These methods have enabled a wide range of applications, including human animation~\citep{yuan2025identity,lai2026slotID,tu2025stableanimator,chang2025x} and keyframe interpolation~\citep{guo2025controllable,zhu2025generative}. 
Early works typically adopt an image-to-video paradigm~\citep{blattmann2023svd,wan2025wan}, where a single reference frame serves as the appearance anchor to guide the generation process. To enhance controllability, more recent approaches extend the reference modality from a single image to multiple images or even video sequences. For instance, several methods exploit video frames to transfer motion patterns from source videos~\citep{chang2025x,cheng2025wananimate,huang2025videomage}, while others utilize multiple reference images to define subjects for customized video synthesis~\citep{cai2025omnivcus,chen2025multi,huang2025conceptmaster,fei2025skyreels,huang2025videomage}. In parallel, Slot-ID~\citep{lai2026slotID} investigates video-based references for identity preservation in general video generation, demonstrating the advantage of richer temporal visual cues.
Despite these advances, existing works primarily focus on appearance preservation or motion transfer under visual conditioning. In contrast, our task of talking avatar generation requires synthesizing temporally coherent facial dynamics driven by audio signals.
Although concurrent work~\citep{lai2026slotID} has explored the use of video references for general video synthesis, the role of video reference in audio-driven talking avatar generation remains largely under-explored.

\noindent\textbf{Talking Avatar Generation.} Talking avatar generation aims to synthesize high-fidelity videos with natural lip synchronization and realistic facial dynamics driven by audio, while preserving the identity of a reference avatar.
Early approaches~\citep{ye2023geneface,ye2024mimictalk,peng2025dualtalk,peng2023emotalk,ma2023styletalk} primarily relied on 3D modeling pipelines that render avatars from explicit geometric representations.
With the rapid progress of video diffusion models, more recent methods~\citep{cui2025hallo3,cui2025hallo4,tian2024emo,ma2026playmate2,chen2025hunyuanvideoavatar,gao2025wans2v,gan2025omniavatar,tu2025stableavatar,li2025infinityhuman,meituanlongcatteam2025longcatvideoavatartechnicalreport} have shifted toward audio-conditioned video generation frameworks, where identity is inferred from visual references.
However, these methods typically condition the generation on a single reference image captured or generated in the same scene as the target video. Such a design limits their ability to generalize across environments and makes cross-scene talking avatar generation difficult to achieve in a straightforward manner. Recently, HuMo~\citep{chen2025humo} took an important step in this direction by enabling cross-scene synthesis through a progressive training strategy. Nevertheless, it still relies on a single reference image with limited identity cues, often resulting in unstable identity preservation under scene variations and speech-driven facial dynamics. 

In this paper, we demonstrate the advantages of leveraging video references for high-fidelity talking avatar generation through comprehensive experiments. We present a novel framework that conditions on cross-scene video inputs and adopts a three-stage training strategy to learn robust identity representations across diverse environments. Rather than relying on a single reference image, our method exploits temporally rich visual cues from video references to preserve identity under scene variations and speech-driven dynamics. To our knowledge, this work is the first to systematically explore cross-scene video references for talking avatar generation.
\section{Method}
\begin{figure*}[!t]
  \centering
  \includegraphics[width=\textwidth]{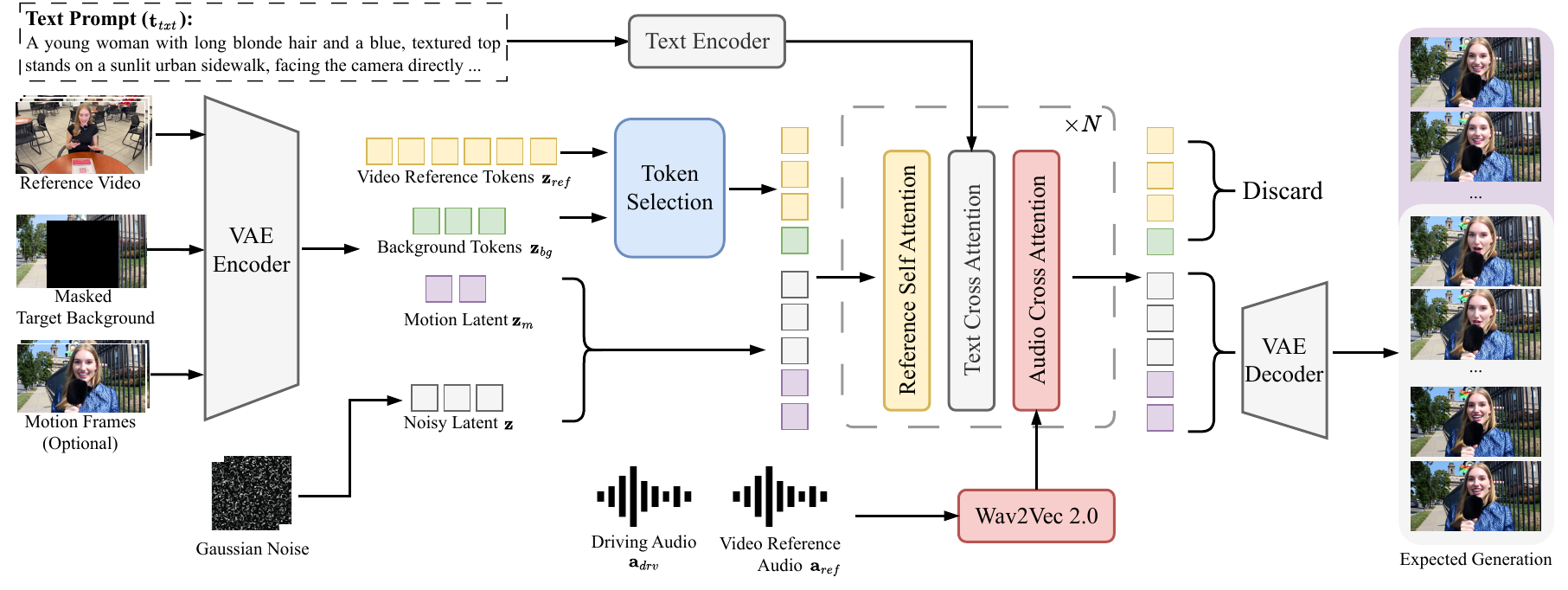}
  \caption{\textbf{Overview of TAVR framework.} Our framework generates high-fidelity talking avatars with customized backgrounds by integrating \textbf{cross-scene video references}. Visual inputs, including the video reference and masked target background, are encoded by the VAE into latents $\mathbf{z}_{ref}$ and $\mathbf{z}_{bg}$, followed by a Token Selection module to reduce computational redundancy. These tokens, alongside an optional motion latent $\mathbf{z}_{m}$ for longer video synthesis, are concatenated with the noisy target latent $\mathbf{z}$ and forwarded through $N$ adapted Transformer blocks. Within each block, a Reference Self-Attention module extends standard self-attention to jointly process target and reference features. Subsequently, two cross-attention modules inject guidance from the text prompt $\mathbf{t}_{txt}$ via text encoder~\citep{chung2023unimax} and audio signals encoded from Wav2Vec 2.0~\citep{baevski2020wav2vec}. Notably, besides the driving audio $\mathbf{a}_{drv}$ for target lip synchronization, the audio module incorporates the corresponding reference audio $\mathbf{a}_{ref}$ to inject explicit audio-visual clues into the reference stream, guiding the network to accurately locate and extract the intrinsic speaking dynamics from the reference tokens. After discarding the auxiliary reference and background tokens, the target generation is decoded by the VAE.}
  \label{fig:main}
\end{figure*}

Given a driving audio signal $\mathbf{a}_{drv}$ and a visual identity reference $\mathbf{R}_{id}$, the talking avatar generation for the target $V$ is formulated as:
\begin{align}
V = \mathcal{G}(\mathbf{a}_{drv}, \mathbf{R}_{id}, \mathbf{C}_{aux}),
\end{align}
where $\mathcal{G}$ denotes a generation model guided by auxiliary control signals $\mathbf{C}_{aux}$, such as text prompts. 
In this work, we aim to address this task by proposing an effective identity-preserving framework, \textbf{TAVR}. As illustrated in Fig.~\ref{fig:main}, this framework integrates temporally rich, cross-scene video references into the generation pipeline to enable high-fidelity synthesis with customized backgrounds. To fully exploit this expanded reference space and ensure robust generalization, TAVR is optimized through a tailored three-stage training strategy.

\subsection{TAVR: Generation from Cross-scene Video References}
TAVR is an identity-preserving framework that integrates cross-scene video references to enable high-fidelity talking avatar generation with customized backgrounds. It fundamentally extends the identity reference space $\mathbf{R}_{id}$ from typical single-image references to multi-frame video sequences of diverse scenes, while expanding the auxiliary control signals $\mathbf{C}_{aux}$ to include both a text prompt $\mathbf{t}_{txt}$ and a masked background image $I_{bg}$ for customized generation. Internally, TAVR contains four key designs: flexible video referencing that dynamically leverages varying lengths of temporal context; 
a token selection module to efficiently condense salient identity cues and background information; 
adapted attention modules, specifically Reference Self-Attention and Audio Cross-Attention, to inject identity cues and essential speech signals with minimal structural modifications; 
and longer video generation enabled by the motion frames strategy to ensure temporal coherence and long-term consistency across successive frames.

\noindent\textbf{Flexible Video Referencing.} 
Unlike previous talking avatar pipelines~\citep{gan2025omniavatar,chen2025humo} that are bottlenecked by a single static image, TAVR expands the identity reference space $\mathbf{R}_{id}$ to accept multi-frame video reference inputs. A key advantage of this design is its flexibility regarding the number of reference frames. By accommodating variable-length video contexts, the model can aggregate dynamic identity information across diverse poses and expressions. To further strengthen the conditioning, we explicitly incorporate the audio stream $\mathbf{a}_{ref}$ from the reference video. This establishes a robust audio-visual prior that guides the network in accurately aligning and extracting the subject's intrinsic speaking dynamics from the visual context.

\noindent\textbf{Token Selection.} 
Video references inherently introduce massive additional visual tokens, which can lead to prohibitive computational overhead. To mitigate the computational cost and encourage the network to focus strictly on salient identity cues, we adopt an explicit token selection module in the latent space. First, spatial-temporal positional embeddings are added to the VAE-encoded reference features to preserve their relative geometry. Next, we derive facial bounding boxes for each reference frame using keypoints extracted via Sapiens~\citep{khirodkar2024sapiens}. We then spatially mask the latent representations, retaining only the tokens within these bounding boxes. Additionally, we exclude the masked tokens of the target background image $I_{bg}$ to further eliminate redundant computation. This efficiently condenses the reference context without sacrificing critical information.

\noindent\textbf{Adapted Attention Modules.} 
In TAVR, we adapt the Wan architecture~\citep{wan2025wan} for talking avatar generation. Specifically, we reformulate the standard self-attention layer into a Reference Self-Attention module and append an Audio Cross-Attention module immediately following the original text cross-attention block.
To effectively inject identity information, the Reference Self-Attention module jointly processes the noisy latent tokens $\mathbf{z}$, the background tokens $\mathbf{z}_{bg}$, and the reference video tokens $\mathbf{z}_{ref}$. For longer video synthesis, an optional motion latent $\mathbf{z}_m$ is prepended to the noisy latent. This augmented sequence, $[\mathbf{z}_m, \mathbf{z}]$ where $[\cdot, \cdot]$ denotes concatenation along the sequence dimension, replaces $\mathbf{z}$ in all subsequent operations to maintain temporal continuity. We extend the standard self-attention module into a two-step operation. First, the target generation and background tokens simultaneously attend to the full combined context. Specifically, defining the query as $Q^R = [\mathbf{z}, \mathbf{z}_{bg}]$ and the keys and values as $K^R = V^R = [\mathbf{z}, \mathbf{z}_{bg}, \mathbf{z}_{ref}]$, this step is formulated as:
\begin{align}
& O^R = \mathrm{ReferenceSelfAttn}(Q^R,K^R,V^R).
\label{eq:ref_attn}
\end{align}
This operation effectively injects rich identity cues into the target generation $\mathbf{z}$ while enabling self-attention-style feature extraction for both $\mathbf{z}$ and $\mathbf{z}_{bg}$. Second, to maintain the structural integrity of the reference representations without being corrupted by the noisy generation stream, we independently reuse the same attention layer to perform a standard self-attention operation exclusively on the reference tokens $\mathbf{z}_{ref}$.
Next, to temporally align the generated facial dynamics with the speech signals, we append an Audio Cross-Attention module at the end of the attention structure. We employ a frame-wise strategy to establish precise spatial-temporal correspondence. 
Given the driving audio features $\mathbf{a}_{drv} \in \mathbb{R}^{T \times L \times d}$ and reference audio $\mathbf{a}_{ref} \in \mathbb{R}^{T_{ref} \times L \times d}$, where $L$ denotes the number of audio tokens per frame, we first reshape the visual tokens to isolate their spatial dimensions, yielding $\mathbf{z} \in \mathbb{R}^{T \times HW \times d}$ and $\mathbf{z}_{ref} \in \mathbb{R}^{T_{ref} \times HW \times d}$. The cross-attention mechanism is then executed by concatenating the visual and audio features along their respective temporal axes. The frame-wise audio cross-attention operation to produce the audio-enhanced visual features $O^{A}$ is formulated as:
\begin{align}
& Q^{A} = [\mathbf{z}, \mathbf{z}_{ref}], \\
& K^{A} = V^{A} = [\mathbf{a}_{drv}, \mathbf{a}_{ref}], \\
& O^{A} = \mathrm{AudioCrossAttn}(Q^{A}, K^{A}, V^{A}),
\label{eq:audio_attn}
\end{align}
where $Q^{A}, O^{A} \in \mathbb{R}^{(T+T_{ref}) \times HW \times d}$ and $K^{A}, V^{A} \in \mathbb{R}^{(T+T_{ref}) \times L \times d}$ denote the query, output, and key, value pairs, respectively.
Crucially, this joint cross-attention mechanism serves two distinct, highly synergistic purposes. On one hand, injecting the driving audio $\mathbf{a}_{drv}$ into the noisy tokens $\mathbf{z}$ precisely dictates the target talking appearance and ensures accurate lip synchronization. On the other hand, injecting the reference audio $\mathbf{a}_{ref}$ into the reference tokens $\mathbf{z}_{ref}$ establishes a strong temporal audio-visual correspondence within the reference stream, providing explicit clues for better reference matching.

\noindent\textbf{Longer Video Generation.} 
By default, our backbone foundational model~\citep{wan2025wan} generates a single clip around 3 seconds. To support the generation of longer talking avatar videos, TAVR employs the motion frames strategy~\citep{xu2024hallo,meituanlongcatteam2025longcatvideoavatartechnicalreport}. Specifically, to ensure smooth dynamics transitions between generation windows, we extract the final two temporal latent from the previously generated clip to serve as motion priors in the latent space. The motion latent is temporally concatenated with the noisy latent sequence of the current clip to guide continuous temporal dynamics. Notably, to maintain long-term identity consistency and explicitly mitigate the accumulation of generation errors over time, we additionally introduce a global appearance anchor. We achieve this by replacing the masked background tokens in the current generation step with the encoded latent representation of the initial frame from the very first generated clip. We provide demo videos for longer video generation on the project page.

\subsection{Three-stage Training Strategy}
\label{sec:3stage}
Synthesizing avatars from cross-scene video references introduces a pronounced domain gap, as the model must aggregate identity cues from frames of varying environments, illuminations, and poses. To bridge this gap and establish robust identity preservation, TAVR employs a progressive three-stage training strategy. Specifically, the training pipeline advances from foundational same-scene appearance copying, through cross-scene domain adaptation, and concludes with task-specific reinforcement learning to explicitly maximize identity fidelity.

\noindent\textbf{Stage 1: Same-Scene Video Pretraining.}
To endow the model with the foundational ability to copy appearance from the reference video and leverage large-scale, unpaired video datasets, we initialize the training of TAVR using intra-scene clips. Employing the flow matching strategy~\citep{lipmanflowmatching,esser2024rectifiedflow} for generative modeling, we construct the optimal transport path as $\mathbf{z}_t = (1-t)\mathbf{z}_0 + t\bm{\epsilon}$, where $\mathbf{z}_0$ represents the clean data latent, $t \in [0,1]$ is the timestep, and $\bm{\epsilon} \sim \mathcal{N}(0, I)$ is standard Gaussian noise. The TAVR network $f_\theta$ is supervised to predict the velocity field target $\mathbf{y} = \bm{\epsilon} - \mathbf{z}_0$ using the following objective:
\begin{align}
  & \mathcal{L}_\mathrm{MSE} = \mathbb{E}_{\mathbf{z}_0,\bm{\epsilon}\sim\mathcal{N}(0,I)}[||f_\theta(\mathbf{a}_{drv}, \mathbf{R}_{id}, \mathbf{C}_{aux},\mathbf{z}_t,t)-\mathbf{y}||_2^2],
  \label{eq:same_scene_pretraining}
\end{align}

\noindent\textbf{Stage 2: Cross-Scene Video Fine-tuning.}
The foundational appearance copying ability acquired during pretraining is insufficient for robust cross-scene synthesis. Constrained by this na\"{i}ve mechanism, the model struggles with new scene adaptation, often resulting in inaccurate facial illumination and unnatural visual composites within customized backgrounds. To bridge this domain gap, we explicitly fine-tune TAVR on video pairs featuring the same identity across distinct backgrounds. In practice, the video reference and the target generation sequence are independently sampled from different videos within the same video pair. During this stage, the network is optimized using the identical flow matching objective $\mathcal{L}_\mathrm{MSE}$ defined in Stage 1. By altering the data distribution while maintaining the reconstruction objective, we compel the model to learn genuine identity aggregation and domain adaptation rather than superficially copying.

\noindent\textbf{Stage 3: Task-Specific Reinforcement Learning.}
To explicitly maximize identity fidelity, we introduce a task-specific reinforcement learning stage based on Direct Preference Optimization (DPO)~\citep{rafailov2023direct}. Following established DPO protocols~\citep{liu2025videodpo,wallace2024diffusiondpo,liu2025videoalign}, we construct preference datasets $\mathcal{D}=\{(\mathbf{c}, \mathbf{z}_0^w, \mathbf{z}_0^l)\}$. Here, $\mathbf{z}_0^w$ represents the winning generation preferred over the losing generation $\mathbf{z}_0^l$, conditioned on the unified context $\mathbf{c} = \{\mathbf{a}_{drv}, \mathbf{R}_{id}, \mathbf{C}_{aux}\}$. To construct these pairs, we employ ArcFace~\citep{deng2019arcface} to measure the identity similarity between the generated and ground-truth videos, designating the sample with the higher similarity score as the winner. 
As our preference reward targets foreground identity, calculating loss over the explicitly provided background $I_{bg}$ is redundant and can introduce spatial noise into the identity learning process. Therefore, we adopt a spatially masked DPO objective to force the network to focus exclusively on the foreground for preserving dynamic identity.
By applying a binary foreground mask $\mathbf{m}$ derived from the background input, we restrict the preference optimization to the avatar region. The task-specific DPO objective is formulated as:
\begin{align}
    & \Delta^* = \|\mathbf{m} \odot(\mathbf{y}^* - f_\theta(\mathbf{c},\mathbf{z}_t^*, t))\|^2 - \|\mathbf{m}\odot(\mathbf{y}^* - f_{\text{ref}}(\mathbf{c},\mathbf{z}_t^*, t))\|^2 \nonumber\\
    & \mathcal{L}_{DPO} = -\mathbb{E} [ \log \sigma ( -\frac{\beta}{2} ( \Delta^w -\Delta^l ) ) ]
\end{align}
where $\odot$ is the Hadamard product, $f_{\text{ref}}$ is the frozen reference model initialized from Stage 2, and $\beta$ is a hyperparameter controlling KL regularization. Finally, to prevent the model from suffering severe distribution shift, we anchor the DPO objective with the original flow matching loss ($\mathcal{L}_{\text{MSE}}$). The overall reinforcement learning objective is thus formulated as $\mathcal{L}_\mathrm{RL}=\lambda_\mathrm{MSE}\mathcal{L}_\mathrm{MSE}+\lambda_\mathrm{DPO}\mathcal{L}_\mathrm{DPO},$ where $\lambda_{\text{MSE}}$ and $\lambda_{\text{DPO}}$ are the respective weighting coefficients.

\section{Experiment}

We present quantitative and qualitative results in Sec.~\ref{sec:flex_videoref} to analyze the function of video reference regarding the number of frames. We compare our method against the state-of-the-art methods in Sec.~\ref{sec:bm}. Finally, Sec.~\ref{sec:ablation} provides a detailed ablation study to justify our framework design and training strategy.

\noindent\textbf{Implementation Details.} 
TAVR is built upon the Wan2.1-T2V-14B~\citep{wan2025wan} backbone and optimized across 32 high-performance 80GB GPUs. During data preprocessing, both the reference videos and target clips are cropped to a spatial resolution of $480\times896$, centered around the subject using human keypoints extracted via Sapiens~\citep{khirodkar2024sapiens}. To explicitly instill the scalable video referencing capability, the target generation sequence is consistently fixed at 81 frames, whereas the reference video length is dynamically sampled between 12 and 20 frames. We provide more details in the supplementary materials.

\noindent\textbf{Benchmark Dataset.}
Existing talking avatar benchmarks predominantly feature single-image reference clips~\citep{chen2025humo,zhang2021hdtf,meng2025echomimicv2}, lacking the cross-scene video pairs required to rigorously evaluate this task. To address this gap, we curate a novel benchmark dataset comprising 158 high-quality cross-scene video pairs derived from TalkVid~\citep{chen2025talkvid}, where each video is of 5-second duration at 25 fps. Specifically, our data processing pipeline operates as follows: First, we group the videos into identity-matched pairs. To ensure strict facial consistency, we enforce a high identity similarity threshold using ArcFace~\citep{deng2019arcface}, explicitly discarding sub-optimal or degraded pairs. Next, to ensure a pronounced cross-scene domain gap, we mask out the human subjects using YOLO11~\citep{Jocher_Ultralytics_YOLO_2023} and compute the PSNR exclusively on the valid background regions between the paired videos. For each unique identity, we retain only the single video pair that yields the lowest background PSNR, effectively ensuring the greatest discrepancy. Finally, we conduct rigorous manual verification to eliminate sequences with abrupt cuts or artifacts, preserving only those with smooth temporal coherence and natural facial dynamics. During evaluation, we systematically designate one video from each pair as the reference and the other as the target. From the target video, we extract the background, the driving audio, and the text prompt. Meanwhile, the reference video supplies the cross-scene visual frames along with their corresponding reference audio. This curated benchmark will be released to facilitate future research in video-referenced talking avatar generation.

\noindent\textbf{Metrics.}
We evaluate the generated avatars across three primary dimensions: identity preservation, lip synchronization, and overall video quality. For identity evaluation, we measure the video-level similarity of each generated video against both its reference and target videos by calculating the frame-wise cosine similarity of their ArcFace~\citep{deng2019arcface} features. Crucially, to ensure this metric remains robust to variations in head pose and facial expressions, we employ a Chamfer similarity formulation for all video-to-video measurements reported in this paper. This approach naturally aligns the most similar frames across the two sequences, effectively filtering out pose-induced noise. 
To quantify lip synchronization, we adopt the standard Sync-C and Sync-D metrics derived from SyncNet~\citep{Chung16a}. Following HuMo~\citep{chen2025humo}, we assess general generation quality using a reward model~\citep{liu2025videoalign} trained on annotated human preference data. This model provides comprehensive evaluations across three distinct perspectives: Visual Quality (VQ), Motion Quality (MQ), and Text Alignment (TA).

\subsection{Flexibility of Video References}
\label{sec:flex_videoref}

\noindent\textbf{Setup.}
To evaluate the robustness and flexibility of our video referencing mechanism, we analyze TAVR's performance across varying lengths of reference context, spanning from 12 to 48 frames. To maintain computational efficiency, this specific ablation is conducted on a representative subset of 50 randomly selected pairs from our benchmark. We compute the full suite of evaluation metrics to provide a comprehensive assessment of how reference length influences identity preservation, lip synchronization, and general quality.

\begin{figure*}[!t]
  \centering
  \includegraphics[width=\textwidth]{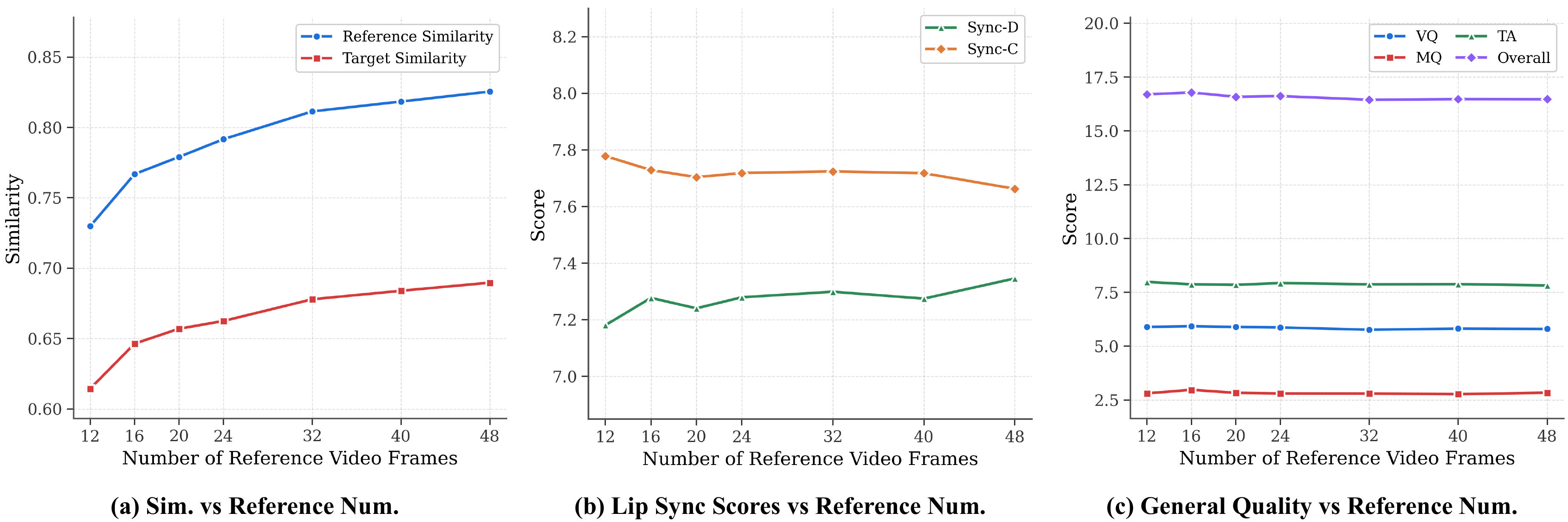}
  \caption{\textbf{Performance with reference number changes.} (a) \textbf{Identity Similarity}: Increasing the reference context leads to a continuous improvement in similarity scores. (b) \textbf{Lip Synchronization}: Sync-D and Sync-C scores remain stable, demonstrating that TAVR maintains precise lip-sync accuracy as the reference length increases. (c) \textbf{General Quality}: Visual quality metrics show negligible fluctuations, proving that the gain in identity fidelity does not compromise generation stability. Notably, TAVR generalizes effectively to 48 reference frames despite its restricted training context of fewer than 20 frames.
  }
  \label{fig:nv_plot}
\end{figure*}

\noindent\textbf{Results.} 
As illustrated in Figure~\ref{fig:nv_plot}, expanding the reference context enhances identity preservation while maintaining highly stable lip synchronization and general video quality. Specifically, Figure~\ref{fig:nv_plot}(a) demonstrates a substantial and continuous improvement in identity similarity as the number of reference frames increases. Notably, while trained on less than 20 frames, TAVR robustly scales to 48 frames. Furthermore, as generation is directly conditioned on the reference sequence, the resulting identity unsurprisingly exhibits higher similarity to the reference than the target.
Crucially, this significant gain in identity fidelity does not compromise other essential generation capabilities. As shown in Figure~\ref{fig:nv_plot}(b) and Figure~\ref{fig:nv_plot}(c), both lip synchronization and general generation quality metrics remain consistently high and stable with the reference increase. While there is a marginal decrease in these scores at the upper bound of reference frames, this behavior is theoretically expected. As the reference visual context expands, the attention mechanism naturally allocates a larger distribution of its capacity toward structural identity preservation. Overall, these results conclusively demonstrate the flexibility in leveraging extended video contexts, proving that longer reference sequences dramatically improve identity preservation with insignificant trade-offs in general generation quality.

\begin{figure*}[!t]
  \centering
  \includegraphics[width=\textwidth]{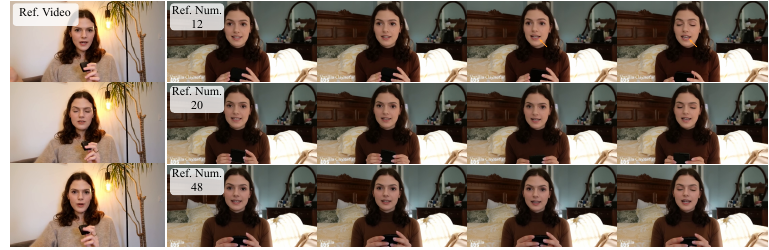}
  \caption{\textbf{Visual comparison across varying reference lengths.} The leftmost column displays the input reference video, followed by the corresponding generation results using different numbers of reference frames. \textcolor[HTML]{FF9933}{Orange} arrows highlight the artifacts in the teeth region of the 12-frame variant.}
  \label{fig:nv_res}
\end{figure*}

\noindent\textbf{Visualizations.} 
Corroborating our quantitative findings, Figure~\ref{fig:nv_res} illustrates that extending the reference context visibly improves identity fidelity. Specifically, relying on a shorter 12-frame context yields inaccurate facial contours and forces the model to hallucinate the occluded teeth region due to missing inner-mouth priors. By scaling the context to 48 frames, the sequence encompasses varied speaking moments, providing the network with the explicit visual cues to preserve the identity accurately. We observe that TAVR achieves this robust identity preservation while ensuring smooth temporal consistency.

\begin{table}[t]
\caption{\textbf{Quantitative comparison with state-of-the-art methods.} }
\small
\label{tab:benchmark}
\centering
\setlength\tabcolsep{3pt}
\resizebox{\textwidth}{!}{
\begin{tabular}{clcccccccc}
\hline
\multirow{2}{*}{Identity Ref.} & \multirow{2}{*}{Method} & \multicolumn{2}{c}{Identity} & \multicolumn{2}{c}{Lip Sync} & \multicolumn{4}{c}{General Generation Quality} \\
\cmidrule(lr){3-4} \cmidrule(lr){5-6} \cmidrule(lr){7-10}
& & ID\textsubscript{ref}$\uparrow$ & ID\textsubscript{target}$\uparrow$ & Sync-C$\uparrow$ & Sync-D$\downarrow$ & VQ$\uparrow$ & MQ$\uparrow$ & TA$\uparrow$ & Overall$\uparrow$ \\
\midrule
\multirow{5}{*}{\shortstack[c]{Cross-scene\\Image}} & StableAvatar~\cite{tu2025stableavatar} & 0.75 & 0.61 & 2.99 & 11.96 & 4.19 & 1.42 & 4.02 & 9.63 \\
& EchoMimic v3~\cite{meng2025echomimicv3} & 0.81 & 0.65 & 4.59 & 10.18 & 4.83 & 2.38 & 3.69 & 10.90 \\
& OmniAvatar~\cite{gan2025omniavatar} & 0.79 & 0.65 & 7.51 & 7.85 & 3.63 & 1.82 & 4.94 & 10.39 \\
& LongCat-Video-Avatar~\cite{meituanlongcatteam2025longcatvideoavatartechnicalreport} & 0.81 & 0.66 & 7.06 & 7.82 & 4.02 & 1.29 & 5.04 & 10.35 \\
& HuMo~\cite{chen2025humo} & 0.73 & 0.62 & 7.12 & 8.45 & 4.43 & 2.03 & 7.67 & 14.13 \\
\midrule
\multirow{5}{*}{\shortstack[c]{Edited\\Image}} & StableAvatar~\cite{tu2025stableavatar} & 0.65 & 0.55 & 2.95 & 11.98 & 3.93 & 1.42 & 7.29 & 12.64 \\
& EchoMimicV3~\cite{meng2025echomimicv3} & 0.69 & 0.58 & 4.66 & 10.16 & 4.63 & 2.47 & 7.02 & 14.12 \\
& OmniAvatar~\cite{gan2025omniavatar} & 0.69 & 0.58 & 7.54 & 7.86 & 3.55 & 2.03 & 7.18 & 12.76 \\
& LongCat-Video-Avatar~\cite{meituanlongcatteam2025longcatvideoavatartechnicalreport} & 0.71 & 0.60 & 7.14 & 7.81 & 4.00 & 1.61 & 7.34 & 12.95 \\
& HuMo~\cite{chen2025humo} & 0.69 & 0.59 & 6.96 & 8.52 & 4.39 & 2.05 & 7.58 & 14.02 \\
\midrule
\multirow{2}{*}{\shortstack[c]{Cross-scene\\Video}} 
& TAVR (20 Ref. Frames) & 0.78 & 0.66 & 7.60 & \textbf{7.41} & \textbf{5.78} & 2.80 & \textbf{7.84} & \textbf{16.42} \\
& TAVR (48 Ref. Frames) & \textbf{0.83} & \textbf{0.69} & \textbf{7.64} & 7.43 & 5.72 & \textbf{2.81} & 7.76 & 16.29 \\
\hline
\end{tabular}
}
\vspace{-10pt}
\end{table}

\begin{figure*}[!t]
  \centering
  \vspace{-10pt}
  \includegraphics[width=0.83\textwidth]{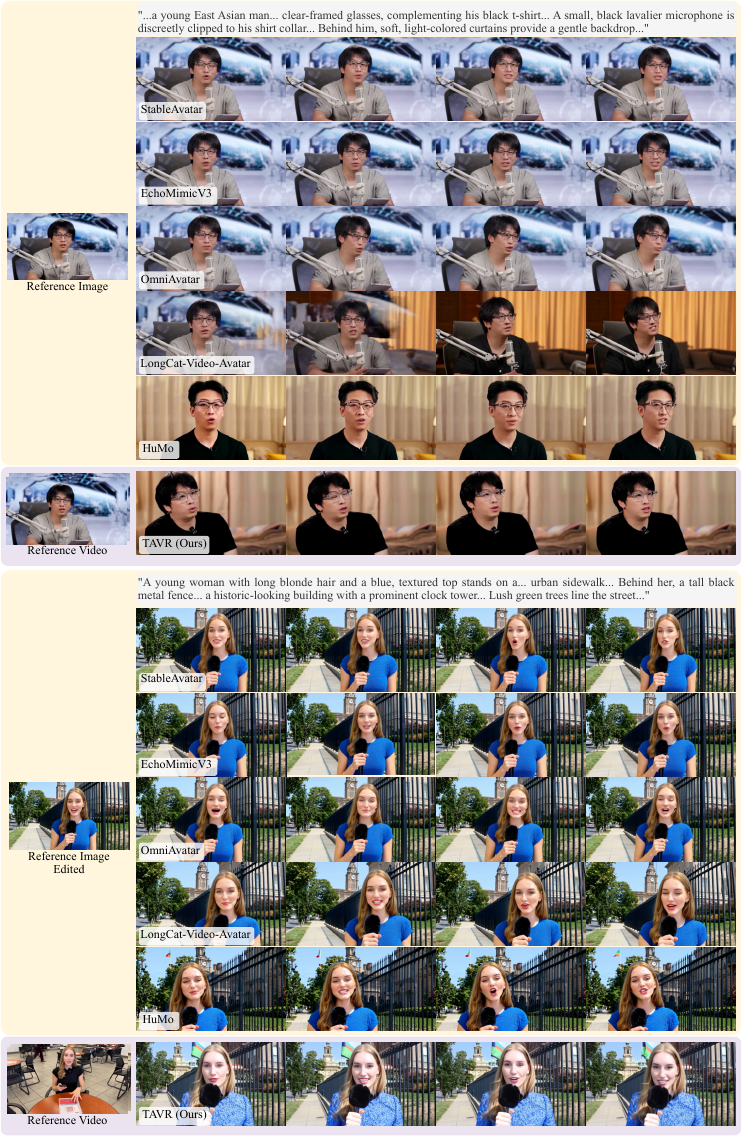}
  \caption{\textbf{Qualitative comparisons with the State-of-The-Art methods.}}
  \label{fig:bm_vis}
\end{figure*}

\subsection{Benchmark Results}
\label{sec:bm}

\noindent\textbf{Setup.} 
We compare TAVR against the state-of-the-art talking avatar generation methods, including StableAvatar~\citep{tu2025stableavatar}, EchoMimicV3~\citep{meng2025echomimicv3}, OmniAvatar~\citep{gan2025omniavatar}, HuMo~\citep{chen2025humo}, and LongCat-Video-Avatar~\citep{meituanlongcatteam2025longcatvideoavatartechnicalreport}. As most of these baselines are inherently designed for the same-scene image referencing, we adapt them for cross-scene evaluation using two distinct testing protocols for the identity reference. In the first paradigm, we provide the raw, unedited \textit{cross-scene image} directly to the baselines as the identity reference. In the second paradigm, we employ a standard two-stage pipeline: the cross-scene reference image is first contextually adapted to the target scene using an advanced image-to-image editor, Qwen-Image-Edit~\citep{wu2025qwen-image}, and the \textit{edited image} is subsequently fed to the baseline. All methods are evaluated across our complete curated benchmark using the full suite of metrics to ensure a rigorous and comprehensive comparison.

\noindent\textbf{Baseline Results Analysis.} 
As detailed in Table~\ref{tab:benchmark}, existing methods severely struggle with the domain gap inherent in cross-scene generation. Given a raw cross-scene image as a reference, most baselines exhibit severe degradation in general video quality, experiencing a particularly sharp drop in Text Alignment (TA). Conversely, while the two-stage pipeline that uses the edited image as the identity reference recovers general video quality in the target scene, it suffers from noticeable identity degradation compared to the direct reference paradigm. This is primarily due to accumulated errors and structural loss inherent in the sequential image-editing phase.
HuMo~\citep{chen2025humo} is a notable method capable of adapting to direct cross-scene references. While it maintains relatively high quality scores using a cross-scene image as the identity reference compared to other baseline methods, its identity scores are noticeably lower. This clearly indicates a fundamental trade-off between cross-scene adaptation and identity preservation.

\noindent\textbf{TAVR Results Analysis.} 
We report the quantitative results of our TAVR framework against the previous methods in Table~\ref{tab:benchmark}. Our method delivers a comprehensive state-of-the-art performance on cross-scene talking avatar generation across all metrics. 
Unlike single-image baselines like HuMo, our approach achieves substantially higher identity fidelity while maintaining top-tier general generation quality.
For instance, our 48-frame variant achieves an $\text{ID}_{\text{target}}$ of 0.69, outperforming the best baselines. Furthermore, TAVR dominates in lip synchronization with a 7.64 Sync-C score. For general generation quality, every variant of our model yields an overall quality score exceeding 16.29, representing a leap over the closest competitor, HuMo, at 14.13.

\noindent\textbf{Visualizations.}
In Figure~\ref{fig:bm_vis}, we present a qualitative comparison between our TAVR framework and state-of-the-art methods. The top two rows depict baselines conditioned on the raw cross-scene image, while the bottom two rows show baselines using the pre-edited reference image adapted to the target scene. Across both settings, TAVR leverages its cross-scene video reference to deliver the highest-fidelity talking avatars with robust identity preservation. Specifically, in the direct cross-scene setting (top case), our method seamlessly generates the avatar within the target background according to the text description, avoiding the unnatural transitions seen in other methods while preserving a noticeably more accurate identity than HuMo. Furthermore, in the two-stage pipeline setting (bottom case), our native video-conditioning naturally avoids the severe error accumulation inherent in sequential image editing. As a result, TAVR generates avatars with significantly better identity than the baselines (e.g., the accurate silhouette in the bottom case).

\subsection{Ablation Study}
\label{sec:ablation}
\begin{table}[t]
\caption{\textbf{Ablation study on training stages.}}
\small
\label{tab:ablation_stage}
\centering
\setlength\tabcolsep{4pt}
\resizebox{0.95\textwidth}{!}{
\begin{tabular}{ccccccccccc}
\hline
\multicolumn{3}{c}{Training Stage} & \multicolumn{2}{c}{Identity} & \multicolumn{2}{c}{Lip Sync} & \multicolumn{4}{c}{Generation Quality} \\
\cmidrule(lr){1-3} \cmidrule(lr){4-5} \cmidrule(lr){6-7} \cmidrule(lr){8-11}
Stage 1 & Stage 2 & Stage 3 & ID\textsubscript{ref}$\uparrow$ & ID\textsubscript{target}$\uparrow$ & Sync-C$\uparrow$ & Sync-D$\downarrow$ & VQ$\uparrow$ & MQ$\uparrow$ & TA$\uparrow$ & Overall$\uparrow$ \\
\midrule
\checkmark & & & \textbf{0.79} & 0.65 & 7.58 & 7.44 & 5.77 & 2.71 & 7.24 & 15.72 \\
& \checkmark & & 0.67 & 0.57 & 7.29 & 7.70 & \textbf{5.82} & \textbf{2.91} & 7.81 & \textbf{16.54} \\
\checkmark & \checkmark & & 0.74 & 0.63 & 7.57 & 7.43 & 5.65 & 2.75 & \textbf{7.85} & 16.25 \\
\checkmark & \checkmark & \checkmark & 0.78 & \textbf{0.66} & \textbf{7.60} & \textbf{7.41} & 5.78 & 2.80 & 7.84 & 16.42 \\
\hline
\end{tabular}
}
\end{table}

\begin{figure*}[!t]
  \centering
  \includegraphics[width=\textwidth]{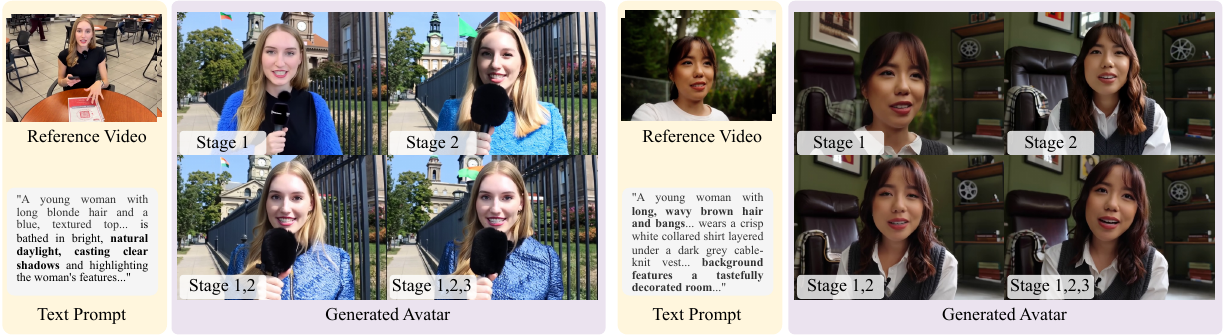}
  \caption{\textbf{Qualitative results of ablated training stages.} We present visualizations for different training stages. Stage 1 denotes same-scene pretraining, Stage 2 applies cross-scene fine-tuning, and Stage 3 integrates task-specific reinforcement learning. Details are provided in Sec.~\ref{sec:3stage}.}
  \label{fig:ablate_vis}
\end{figure*}

We perform an ablation study on our three-stage training pipeline, detailing quantitative results in Table~\ref{tab:ablation_stage} and qualitative comparisons in Figure~\ref{fig:ablate_vis}. Additional ablations are provided in the appendix.

\noindent\textbf{Same-scene Video Pretraining.} 
This `Stage 1' is designed to learn fundamental appearance copying from the reference video, achieving a high ID\textsubscript{ref} of 0.79. Omitting this pretraining, as the comparison between `Stage 2' and `Stage 1,2', results in a 0.06 drop in ID\textsubscript{target} and noticeable identity degradation in the visualizations. These results demonstrate the necessity of same-scene video pretraining for robust identity preservation.

\noindent\textbf{Cross-scene Video Fine-tuning.} 
As the second stage, this `Stage 2' trains TAVR on cross-scene video reference pairs for cross-scene adaptation. This capability is quantitatively supported in Table~\ref{tab:ablation_stage}, where variants incorporating `Stage 2' achieve high TA scores exceeding 7.80. Specifically, omitting this stage and relying solely on `Stage 1' causes a significant TA drop of 0.61 compared to the combined `Stage 1,2' variant.
Qualitatively, as shown in Figure~\ref{fig:ablate_vis}, the copying mechanism of `Stage 1' struggles with environmental changes. In the left example, the model fails to synthesize the specific facial illumination explicitly dictated by the text prompt. In the right example, it rigidly pastes the reference avatar into the new scene, resulting in an unnatural visual composite. These observations strongly highlight the necessity of cross-scene video fine-tuning.

\noindent\textbf{Task-specific Reinforcement Learning.} 
This `Stage 3' is explicitly designed to maximize identity preservation. As shown in Table~\ref{tab:ablation_stage}, integrating this final stage yields absolute improvements of 0.04 and 0.03 on the ID\textsubscript{ref} and ID\textsubscript{target} metrics, respectively. This clearly validates the effectiveness of our reinforcement learning strategy. Additional ablations for this variant are provided in the appendix.

\section{Conclusion}
We propose TAVR, a robust cross-scene talking avatar generation framework powered by our novel video-referencing paradigm and token selection module. TAVR effectively leverages rich temporal identity cues from cross-scene video reference through a comprehensive three-stage training scheme, enabling the generation of highly-fidelity avatars within customized backgrounds. For evaluation, we construct a novel cross-scene talking avatar benchmark comprising 158 high-quality video pairs with pronounced scene differences. Extensive experiments on the benchmark demonstrate that TAVR consistently outperforms previous methods, achieving state-of-the-art performance with unprecedented identity preservation, precise lip synchronization, and top-tier video quality.

{\small
\bibliographystyle{plainnat}
\bibliography{main}

\begin{thebibliography}{51}
\providecommand{\natexlab}[1]{#1}
\providecommand{\url}[1]{\texttt{#1}}
\expandafter\ifx\csname urlstyle\endcsname\relax
  \providecommand{\doi}[1]{doi: #1}\else
  \providecommand{\doi}{doi: \begingroup \urlstyle{rm}\Url}\fi

\bibitem[Baevski et~al.(2020)Baevski, Zhou, Mohamed, and Auli]{baevski2020wav2vec}
Alexei Baevski, Yuhao Zhou, Abdelrahman Mohamed, and Michael Auli.
\newblock wav2vec 2.0: A framework for self-supervised learning of speech representations.
\newblock In \emph{NeurIPS}, 2020.

\bibitem[Blattmann et~al.(2023)Blattmann, Dockhorn, Kulal, Mendelevitch, Kilian, Lorenz, Levi, English, Voleti, Letts, et~al.]{blattmann2023svd}
Andreas Blattmann, Tim Dockhorn, Sumith Kulal, Daniel Mendelevitch, Maciej Kilian, Dominik Lorenz, Yam Levi, Zion English, Vikram Voleti, Adam Letts, et~al.
\newblock Stable video diffusion: Scaling latent video diffusion models to large datasets.
\newblock \emph{arXiv preprint arXiv:2311.15127}, 2023.

\bibitem[Cai et~al.(2025)Cai, Zhang, Chen, Xing, Hu, Zhou, Zhang, Zhang, Kim, Wang, et~al.]{cai2025omnivcus}
Yuanhao Cai, He~Zhang, Xi~Chen, Jinbo Xing, Yiwei Hu, Yuqian Zhou, Kai Zhang, Zhifei Zhang, Soo~Ye Kim, Tianyu Wang, et~al.
\newblock {OmniVCus}: Feedforward subject-driven video customization with multimodal control conditions.
\newblock In \emph{NeurIPS}, 2025.

\bibitem[Chang et~al.(2025)Chang, Xu, Xie, Gao, Kuang, Cai, Zhang, Song, Wang, Shi, et~al.]{chang2025x}
Di~Chang, Hongyi Xu, You Xie, Yipeng Gao, Zhengfei Kuang, Shengqu Cai, Chenxu Zhang, Guoxian Song, Chao Wang, Yichun Shi, et~al.
\newblock {X-Dyna}: Expressive dynamic human image animation.
\newblock In \emph{CVPR}, 2025.

\bibitem[Chen et~al.(2025{\natexlab{a}})Chen, Ma, Liu, Li, Chen, Liu, He, Li, He, and Wu]{chen2025humo}
Liyang Chen, Tianxiang Ma, Jiawei Liu, Bingchuan Li, Zhuowei Chen, Lijie Liu, Xu~He, Gen Li, Qian He, and Zhiyong Wu.
\newblock {HuMo}: Human-centric video generation via collaborative multi-modal conditioning.
\newblock \emph{arXiv preprint arXiv:2509.08519}, 2025{\natexlab{a}}.

\bibitem[Chen et~al.(2025{\natexlab{b}})Chen, Huang, Liu, Ye, Chen, Zhu, Guan, Wang, Chen, Li, et~al.]{chen2025talkvid}
Shunian Chen, Hejin Huang, Yexin Liu, Zihan Ye, Pengcheng Chen, Chenghao Zhu, Michael Guan, Rongsheng Wang, Junying Chen, Guanbin Li, et~al.
\newblock {TalkVid}: A large-scale diversified dataset for audio-driven talking head synthesis.
\newblock \emph{arXiv preprint arXiv:2508.13618}, 2025{\natexlab{b}}.

\bibitem[Chen et~al.(2025{\natexlab{c}})Chen, Siarohin, Menapace, Fang, Lee, Skorokhodov, Aberman, Zhu, Yang, and Tulyakov]{chen2025multi}
Tsai-Shien Chen, Aliaksandr Siarohin, Willi Menapace, Yuwei Fang, Kwot~Sin Lee, Ivan Skorokhodov, Kfir Aberman, Jun-Yan Zhu, Ming-Hsuan Yang, and Sergey Tulyakov.
\newblock Multi-subject open-set personalization in video generation.
\newblock In \emph{CVPR}, 2025{\natexlab{c}}.

\bibitem[Chen et~al.(2025{\natexlab{d}})Chen, Liang, Zhou, Huang, Ma, Tang, Lin, Zhou, and Lu]{chen2025hunyuanvideoavatar}
Yi~Chen, Sen Liang, Zixiang Zhou, Ziyao Huang, Yifeng Ma, Junshu Tang, Qin Lin, Yuan Zhou, and Qinglin Lu.
\newblock {HunyuanVideo-Avatar}: High-fidelity audio-driven human animation for multiple characters.
\newblock \emph{arXiv preprint arXiv:2505.20156}, 2025{\natexlab{d}}.

\bibitem[Cheng et~al.(2025)Cheng, Gao, Hu, Hu, Huang, Ji, Li, Meng, Qi, Qiao, et~al.]{cheng2025wananimate}
Gang Cheng, Xin Gao, Li~Hu, Siqi Hu, Mingyang Huang, Chaonan Ji, Ju~Li, Dechao Meng, Jinwei Qi, Penchong Qiao, et~al.
\newblock {Wan-Animate}: Unified character animation and replacement with holistic replication.
\newblock \emph{arXiv preprint arXiv:2509.14055}, 2025.

\bibitem[Chung et~al.(2023)Chung, Constant, Garcia, Roberts, Tay, Narang, and Firat]{chung2023unimax}
Hyung~Won Chung, Noah Constant, Xavier Garcia, Adam Roberts, Yi~Tay, Sharan Narang, and Orhan Firat.
\newblock {UniMax}: Fairer and more effective language sampling for large-scale multilingual pretraining.
\newblock In \emph{ICLR}, 2023.

\bibitem[Chung and Zisserman(2016)]{Chung16a}
J.~S. Chung and A.~Zisserman.
\newblock Out of time: automated lip sync in the wild.
\newblock In \emph{Workshop on Multi-view Lip-reading, ACCV}, 2016.

\bibitem[Cui et~al.(2025{\natexlab{a}})Cui, Chen, Xu, Shang, Chen, Zhan, Dong, Yao, Wang, and Zhu]{cui2025hallo4}
Jiahao Cui, Yan Chen, Mingwang Xu, Hanlin Shang, Yuxuan Chen, Yun Zhan, Zilong Dong, Yao Yao, Jingdong Wang, and Siyu Zhu.
\newblock {Hallo4}: High-fidelity dynamic portrait animation via direct preference optimization and temporal motion modulation.
\newblock \emph{arXiv preprint arXiv:2505.23525}, 2025{\natexlab{a}}.

\bibitem[Cui et~al.(2025{\natexlab{b}})Cui, Li, Zhan, Shang, Cheng, Ma, Mu, Zhou, Wang, and Zhu]{cui2025hallo3}
Jiahao Cui, Hui Li, Yun Zhan, Hanlin Shang, Kaihui Cheng, Yuqi Ma, Shan Mu, Hang Zhou, Jingdong Wang, and Siyu Zhu.
\newblock {Hallo3}: Highly dynamic and realistic portrait image animation with video diffusion transformer.
\newblock In \emph{CVPR}, 2025{\natexlab{b}}.

\bibitem[Deng et~al.(2019)Deng, Guo, Xue, and Zafeiriou]{deng2019arcface}
Jiankang Deng, Jia Guo, Niannan Xue, and Stefanos Zafeiriou.
\newblock {Arcface}: Additive angular margin loss for deep face recognition.
\newblock In \emph{CVPR}, 2019.

\bibitem[Esser et~al.(2024)Esser, Kulal, Blattmann, Entezari, M{\"u}ller, Saini, Levi, Lorenz, Sauer, Boesel, et~al.]{esser2024rectifiedflow}
Patrick Esser, Sumith Kulal, Andreas Blattmann, Rahim Entezari, Jonas M{\"u}ller, Harry Saini, Yam Levi, Dominik Lorenz, Axel Sauer, Frederic Boesel, et~al.
\newblock Scaling rectified flow transformers for high-resolution image synthesis.
\newblock In \emph{ICML}, 2024.

\bibitem[Fei et~al.(2025)Fei, Li, Qiu, Wang, Dou, Wang, Xu, Fan, Chen, Li, et~al.]{fei2025skyreels}
Zhengcong Fei, Debang Li, Di~Qiu, Jiahua Wang, Yikun Dou, Rui Wang, Jingtao Xu, Mingyuan Fan, Guibin Chen, Yang Li, et~al.
\newblock {SkyReels-A2}: Compose anything in video diffusion transformers.
\newblock \emph{arXiv preprint arXiv:2504.02436}, 2025.

\bibitem[Gan et~al.(2025)Gan, Yang, Zhu, Xue, and Hoi]{gan2025omniavatar}
Qijun Gan, Ruizi Yang, Jianke Zhu, Shaofei Xue, and Steven Hoi.
\newblock {OmniAvatar}: Efficient audio-driven avatar video generation with adaptive body animation.
\newblock \emph{arXiv preprint arXiv:2506.18866}, 2025.

\bibitem[Gao et~al.(2025)Gao, Hu, Hu, Huang, Ji, Meng, Qi, Qiao, Shen, Song, et~al.]{gao2025wans2v}
Xin Gao, Li~Hu, Siqi Hu, Mingyang Huang, Chaonan Ji, Dechao Meng, Jinwei Qi, Penchong Qiao, Zhen Shen, Yafei Song, et~al.
\newblock {Wan-S2V}: Audio-driven cinematic video generation.
\newblock \emph{arXiv preprint arXiv:2508.18621}, 2025.

\bibitem[Guo et~al.(2025)Guo, Wu, Cai, Li, and Loy]{guo2025controllable}
Zujin Guo, Size Wu, Zhongang Cai, Wei Li, and Chen~Change Loy.
\newblock Controllable human-centric keyframe interpolation with generative prior.
\newblock In \emph{NeurIPS}, 2025.

\bibitem[Ho and Salimans(2021)]{ho2021classifier}
Jonathan Ho and Tim Salimans.
\newblock Classifier-free diffusion guidance.
\newblock In \emph{NeurIPS Workshop on Deep Generative Models and Downstream Applications}, 2021.

\bibitem[Huang et~al.(2025{\natexlab{a}})Huang, Wu, Chung, Chang, Yang, and Wang]{huang2025videomage}
Chi-Pin Huang, Yen-Siang Wu, Hung-Kai Chung, Kai-Po Chang, Fu-En Yang, and Yu-Chiang~Frank Wang.
\newblock {VideoMage}: Multi-subject and motion customization of text-to-video diffusion models.
\newblock In \emph{CVPR}, 2025{\natexlab{a}}.

\bibitem[Huang et~al.(2025{\natexlab{b}})Huang, Yuan, Liu, Wang, Wang, Zhang, Wan, Zhang, and Gai]{huang2025conceptmaster}
Yuzhou Huang, Ziyang Yuan, Quande Liu, Qiulin Wang, Xintao Wang, Ruimao Zhang, Pengfei Wan, Di~Zhang, and Kun Gai.
\newblock {ConceptMaster}: Multi-concept video customization on diffusion transformer models without test-time tuning.
\newblock \emph{arXiv preprint arXiv:2501.04698}, 2025{\natexlab{b}}.

\bibitem[Jocher et~al.(2023)Jocher, Qiu, and Chaurasia]{Jocher_Ultralytics_YOLO_2023}
Glenn Jocher, Jing Qiu, and Ayush Chaurasia.
\newblock {Ultralytics YOLO}, January 2023.
\newblock URL \url{https://github.com/ultralytics/ultralytics}.

\bibitem[Khirodkar et~al.(2024)Khirodkar, Bagautdinov, Martinez, Zhaoen, James, Selednik, Anderson, and Saito]{khirodkar2024sapiens}
Rawal Khirodkar, Timur Bagautdinov, Julieta Martinez, Su~Zhaoen, Austin James, Peter Selednik, Stuart Anderson, and Shunsuke Saito.
\newblock {Sapiens}: Foundation for human vision models.
\newblock In \emph{ECCV}, 2024.

\bibitem[Lai et~al.(2026)Lai, Wang, Zhou, and Shao]{lai2026slotID}
Yixuan Lai, He~Wang, Kun Zhou, and Tianjia Shao.
\newblock {Slot-ID}: Identity-preserving video generation from reference videos via slot-based temporal identity encoding.
\newblock \emph{arXiv preprint arXiv:2601.01352}, 2026.

\bibitem[Li et~al.(2025{\natexlab{a}})Li, Xie, Ren, Gan, Zhang, Kong, Yin, Peng, and Yuan]{li2025infinityhuman}
Xiaodi Li, Pan Xie, Yi~Ren, Qijun Gan, Chen Zhang, Fangyuan Kong, Xiang Yin, Bingyue Peng, and Zehuan Yuan.
\newblock {InfinityHuman}: Towards long-term audio-driven human animation.
\newblock \emph{arXiv preprint arXiv:2508.20210}, 2025{\natexlab{a}}.

\bibitem[Li et~al.(2025{\natexlab{b}})Li, Qian, Su, Diao, Xia, Liu, Yang, Zhang, and Yuan]{li2025bindweave}
Zhaoyang Li, Dongjun Qian, Kai Su, Qishuai Diao, Xiangyang Xia, Chang Liu, Wenfei Yang, Tianzhu Zhang, and Zehuan Yuan.
\newblock {BindWeave}: Subject-consistent video generation via cross-modal integration.
\newblock In \emph{ICLR}, 2025{\natexlab{b}}.

\bibitem[Lipman et~al.(2022)Lipman, Chen, Ben-Hamu, Nickel, and Le]{lipmanflowmatching}
Yaron Lipman, Ricky~TQ Chen, Heli Ben-Hamu, Maximilian Nickel, and Matthew Le.
\newblock Flow matching for generative modeling.
\newblock In \emph{ICLR}, 2022.

\bibitem[Liu et~al.(2025{\natexlab{a}})Liu, Liu, Liang, Yuan, Liu, Zheng, Wu, Wang, Xia, Wang, et~al.]{liu2025videoalign}
Jie Liu, Gongye Liu, Jiajun Liang, Ziyang Yuan, Xiaokun Liu, Mingwu Zheng, Xiele Wu, Qiulin Wang, Menghan Xia, Xintao Wang, et~al.
\newblock Improving video generation with human feedback.
\newblock In \emph{NeurIPS}, 2025{\natexlab{a}}.

\bibitem[Liu et~al.(2025{\natexlab{b}})Liu, Wu, Zheng, Wei, He, Pi, and Chen]{liu2025videodpo}
Runtao Liu, Haoyu Wu, Ziqiang Zheng, Chen Wei, Yingqing He, Renjie Pi, and Qifeng Chen.
\newblock {VideoDPO}: Omni-preference alignment for video diffusion generation.
\newblock In \emph{CVPR}, 2025{\natexlab{b}}.

\bibitem[Ma et~al.(2026)Ma, Huang, Cai, Guan, Zheng, Zhao, Zhang, and Zhang]{ma2026playmate2}
Xingpei Ma, Shenneng Huang, Jiaran Cai, Yuansheng Guan, Shen Zheng, Hanfeng Zhao, Qiang Zhang, and Shunsi Zhang.
\newblock {Playmate2}: Training-free multi-character audio-driven animation via diffusion transformer with reward feedback.
\newblock In \emph{AAAI}, 2026.

\bibitem[Ma et~al.(2023)Ma, Wang, Hu, Fan, Lv, Ding, Deng, and Yu]{ma2023styletalk}
Yifeng Ma, Suzhen Wang, Zhipeng Hu, Changjie Fan, Tangjie Lv, Yu~Ding, Zhidong Deng, and Xin Yu.
\newblock {Styletalk}: One-shot talking head generation with controllable speaking styles.
\newblock In \emph{AAAI}, 2023.

\bibitem[Meng et~al.(2025)Meng, Zhang, Li, and Ma]{meng2025echomimicv2}
Rang Meng, Xingyu Zhang, Yuming Li, and Chenguang Ma.
\newblock {EchomimicV2}: Towards striking, simplified, and semi-body human animation.
\newblock In \emph{CVPR}, 2025.

\bibitem[Meng et~al.(2026)Meng, Wang, Wu, Zheng, Li, and Ma]{meng2025echomimicv3}
Rang Meng, Yan Wang, Weipeng Wu, Ruobing Zheng, Yuming Li, and Chenguang Ma.
\newblock {EchomimicV3}: 1.3 b parameters are all you need for unified multi-modal and multi-task human animation.
\newblock In \emph{AAAI}, 2026.

\bibitem[Peng et~al.(2023)Peng, Wu, Song, Xu, Zhu, He, Liu, and Fan]{peng2023emotalk}
Ziqiao Peng, Haoyu Wu, Zhenbo Song, Hao Xu, Xiangyu Zhu, Jun He, Hongyan Liu, and Zhaoxin Fan.
\newblock {EmoTalk}: Speech-driven emotional disentanglement for 3d face animation.
\newblock In \emph{ICCV}, 2023.

\bibitem[Peng et~al.(2025)Peng, Fan, Wu, Wang, Liu, He, and Fan]{peng2025dualtalk}
Ziqiao Peng, Yanbo Fan, Haoyu Wu, Xuan Wang, Hongyan Liu, Jun He, and Zhaoxin Fan.
\newblock {DualTalk}: Dual-speaker interaction for 3d talking head conversations.
\newblock In \emph{CVPR}, 2025.

\bibitem[Rafailov et~al.(2023)Rafailov, Sharma, Mitchell, Manning, Ermon, and Finn]{rafailov2023direct}
Rafael Rafailov, Archit Sharma, Eric Mitchell, Christopher~D Manning, Stefano Ermon, and Chelsea Finn.
\newblock Direct preference optimization: Your language model is secretly a reward model.
\newblock In \emph{NeurIPS}, 2023.

\bibitem[Team(2025)]{meituanlongcatteam2025longcatvideoavatartechnicalreport}
Meituan~LongCat Team.
\newblock Longcat-video-avatar technical report.
\newblock \url{https://github.com/meituan-longcat/LongCat-Video/blob/main/assets/LongCat-Video-Avatar-Tech-Report.pdf}, 2025.

\bibitem[Tian et~al.(2024)Tian, Wang, Zhang, and Bo]{tian2024emo}
Linrui Tian, Qi~Wang, Bang Zhang, and Liefeng Bo.
\newblock {EMO}: Emote portrait alive generating expressive portrait videos with audio2video diffusion model under weak conditions.
\newblock In \emph{ECCV}, 2024.

\bibitem[Tu et~al.(2025{\natexlab{a}})Tu, Pan, Huang, Han, Xing, Dai, Luo, Wu, and Jiang]{tu2025stableavatar}
Shuyuan Tu, Yueming Pan, Yinming Huang, Xintong Han, Zhen Xing, Qi~Dai, Chong Luo, Zuxuan Wu, and Yu-Gang Jiang.
\newblock {StableAvatar}: Infinite-length audio-driven avatar video generation.
\newblock \emph{arXiv preprint arXiv:2508.08248}, 2025{\natexlab{a}}.

\bibitem[Tu et~al.(2025{\natexlab{b}})Tu, Xing, Han, Cheng, Dai, Luo, and Wu]{tu2025stableanimator}
Shuyuan Tu, Zhen Xing, Xintong Han, Zhi-Qi Cheng, Qi~Dai, Chong Luo, and Zuxuan Wu.
\newblock {StableAnimator}: High-quality identity-preserving human image animation.
\newblock In \emph{CVPR}, 2025{\natexlab{b}}.

\bibitem[Wallace et~al.(2024)Wallace, Dang, Rafailov, Zhou, Lou, Purushwalkam, Ermon, Xiong, Joty, and Naik]{wallace2024diffusiondpo}
Bram Wallace, Meihua Dang, Rafael Rafailov, Linqi Zhou, Aaron Lou, Senthil Purushwalkam, Stefano Ermon, Caiming Xiong, Shafiq Joty, and Nikhil Naik.
\newblock Diffusion model alignment using direct preference optimization.
\newblock In \emph{CVPR}, 2024.

\bibitem[Wan et~al.(2025)Wan, Wang, Ai, Wen, Mao, Xie, Chen, Yu, Zhao, Yang, et~al.]{wan2025wan}
Team Wan, Ang Wang, Baole Ai, Bin Wen, Chaojie Mao, Chen-Wei Xie, Di~Chen, Feiwu Yu, Haiming Zhao, Jianxiao Yang, et~al.
\newblock {Wan}: Open and advanced large-scale video generative models.
\newblock \emph{arXiv preprint arXiv:2503.20314}, 2025.

\bibitem[Wu et~al.(2025)Wu, Li, Zhou, Lin, Gao, Yan, Yin, Bai, Xu, Chen, et~al.]{wu2025qwen-image}
Chenfei Wu, Jiahao Li, Jingren Zhou, Junyang Lin, Kaiyuan Gao, Kun Yan, Sheng-ming Yin, Shuai Bai, Xiao Xu, Yilei Chen, et~al.
\newblock Qwen-image technical report.
\newblock \emph{arXiv preprint arXiv:2508.02324}, 2025.

\bibitem[Xu et~al.(2024)Xu, Li, Su, Shang, Zhang, Liu, Wang, Yao, and Zhu]{xu2024hallo}
Mingwang Xu, Hui Li, Qingkun Su, Hanlin Shang, Liwei Zhang, Ce~Liu, Jingdong Wang, Yao Yao, and Siyu Zhu.
\newblock {Hallo}: Hierarchical audio-driven visual synthesis for portrait image animation.
\newblock \emph{arXiv preprint arXiv:2406.08801}, 2024.

\bibitem[Ye et~al.(2023)Ye, Jiang, Ren, Liu, He, and Zhao]{ye2023geneface}
Zhenhui Ye, Ziyue Jiang, Yi~Ren, Jinglin Liu, Jinzheng He, and Zhou Zhao.
\newblock {GeneFace}: Generalized and high-fidelity audio-driven 3d talking face synthesis.
\newblock In \emph{ICLR}, 2023.

\bibitem[Ye et~al.(2024)Ye, Zhong, Ren, Jiang, Huang, Huang, Liu, He, Zhang, Wang, et~al.]{ye2024mimictalk}
Zhenhui Ye, Tianyun Zhong, Yi~Ren, Ziyue Jiang, Jiawei Huang, Rongjie Huang, Jinglin Liu, Jinzheng He, Chen Zhang, Zehan Wang, et~al.
\newblock {MimicTalk}: Mimicking a personalized and expressive 3d talking face in minutes.
\newblock \emph{NeurIPS}, 2024.

\bibitem[Yuan et~al.(2025)Yuan, Huang, He, Ge, Shi, Chen, Luo, and Yuan]{yuan2025identity}
Shenghai Yuan, Jinfa Huang, Xianyi He, Yunyang Ge, Yujun Shi, Liuhan Chen, Jiebo Luo, and Li~Yuan.
\newblock Identity-preserving text-to-video generation by frequency decomposition.
\newblock In \emph{CVPR}, 2025.

\bibitem[Zhang et~al.(2021)Zhang, Li, Ding, and Fan]{zhang2021hdtf}
Zhimeng Zhang, Lincheng Li, Yu~Ding, and Changjie Fan.
\newblock Flow-guided one-shot talking face generation with a high-resolution audio-visual dataset.
\newblock In \emph{CVPR}, 2021.

\bibitem[Zhao et~al.(2023)Zhao, Bai, Rao, Zhou, and Lu]{zhao2023unipc}
Wenliang Zhao, Lujia Bai, Yongming Rao, Jie Zhou, and Jiwen Lu.
\newblock Unipc: A unified predictor-corrector framework for fast sampling of diffusion models.
\newblock In \emph{NeurIPS}, 2023.

\bibitem[Zhu et~al.(2025)Zhu, Ren, Wang, Wu, and Zuo]{zhu2025generative}
Tianyi Zhu, Dongwei Ren, Qilong Wang, Xiaohe Wu, and Wangmeng Zuo.
\newblock Generative inbetweening through frame-wise conditions-driven video generation.
\newblock In \emph{CVPR}, 2025.

\end{thebibliography}
}
\clearpage
\appendix
\section{Appendix}

This supplementary document provides further implementation details and additional ablation studies referenced in the main text. Specifically, Sec.~\ref{sec:add_impl} provides additional implementation details. 
We also provide additional qualitative results in Sec.~\ref{sec:add_qual}. 
In Sec.~\ref{sec:abl_tavr}, we analyze the core components of the TAVR framework, focusing on the reference audio conditioning and the Token Selection module. Sec.~\ref{sec:abl_rl} presents targeted ablation studies evaluating our masked DPO settings in reinforcement learning, while Sec.~\ref{sec:limit} discusses the current limitations of our approach. 

\subsection{Additional Implementation Details}
\label{sec:add_impl}
\noindent\textbf{Training.} TAVR is trained with our proposed three-stage training strategy, including same-scene video pretraining, cross-scene video fine-tuning, and task-specific reinforcement learning. During the first two stages, the network is optimized using AdamW with a learning rate of $5\times10^{-6}$ and a 50-step warmup, training for 18k and 15k steps, respectively. For DPO in the task-specific reinforcement learning stage, we curate a highly discriminative dataset of 800 preference pairs, enforcing a strict minimum identity similarity margin of 0.20 between the winning and losing generations. The model is then fine-tuned for 400 steps at a reduced learning rate of $1.5\times10^{-6}$, with the KL regularization hyperparameter set to $\beta=500$. For the objective weights, we configured them as $\lambda_{\text{MSE}}=1$ and $\lambda_{\text{DPO}}=2$, respectively.

\noindent\textbf{Inference.} During inference, we use UniPC~\citep{zhao2023unipc} as the scheduler, with a sampling step of 24. We also leverage the Classifier-Free Guidance (CFG)~\citep{ho2021classifier}, with the guide scales as 1.8 and 5.0 for audio and text guidance, respectively.

\subsection{Additional Qualitative Results}
\label{sec:add_qual}
\begin{figure*}[!t]
  \centering
  \vspace{-10pt}
  \includegraphics[width=0.83\textwidth]{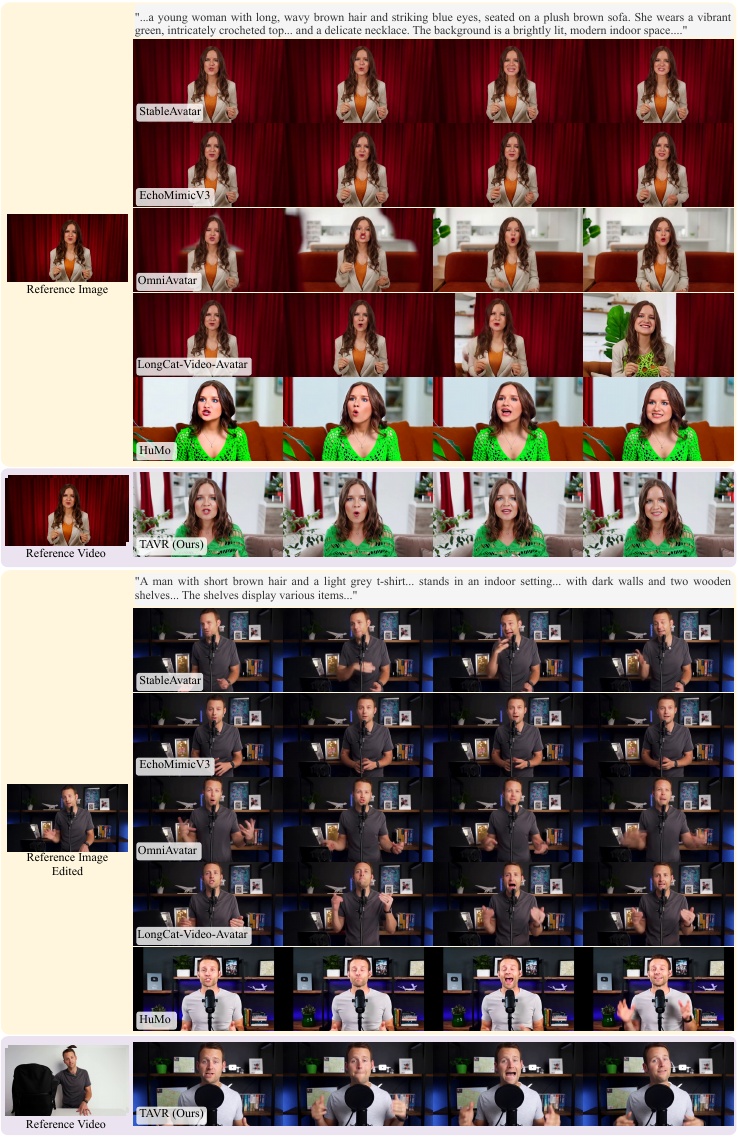}
  \caption{\textbf{Additional qualitative comparisons on the cross-scene benchmark.}}
  \label{fig:add_qual}
\end{figure*}

To further demonstrate the robustness of our approach, we provide extended qualitative comparisons against state-of-the-art methods on our cross-scene benchmark in Figure~\ref{fig:add_qual}. This directly supplements the visual evidence presented in Figure~\ref{fig:bm_vis} of the main text.

\subsection{Ablation on the TAVR framework}
\label{sec:abl_tavr}
We conduct additional experiments to ablate the core architectural designs of our proposed TAVR framework. Specifically, we analyze the critical role of the video reference audio and evaluate the computational efficiency gained by our token selection module. All experiments in this section are conducted using a standard 20-frame video reference.

\noindent\textbf{Video Reference Audio.} 
We incorporate the video reference audio to establish a robust audio-visual prior, which actively guides the network in querying the correct dynamic identity features. To validate its impact, we evaluate a variant without this input, denoted as `w/o Ref. Audio'. As reported in Table~\ref{tab:ablation_audio}, while the model maintains a strong baseline without audio, incorporating the reference audio yields consistent improvements across all metrics. Specifically, the audio reference prior boosts ID\textsubscript{ref} and ID\textsubscript{target} by 0.01 and 0.02, respectively, while simultaneously enhancing lip synchronization with an increase of Sync-C from 7.55 to 7.60. These consistent gains demonstrate that the reference audio provides beneficial guidance for our sequence-based identity referencing, enabling the model to more precisely align and extract talking identity correspondences.

\noindent\textbf{Token Selection Module.}
To mitigate the prohibitive computational cost introduced by massive visual reference tokens, we incorporate an explicit token selection module that actively discards redundant spatial information and isolates salient identity cues. To quantify this efficiency improvement, we compare TAVR against a variant removing this component, denoted as `w/o Token Selection'. 
We evaluate efficiency by reporting the average number of processed tokens and calculating the FLOPs for a single Reference Self-Attention layer. As detailed in Table~\ref{tab:ablation_token}, integrating the token selection module yields a substantial 16.9\% reduction in token count, decreasing from 45,360 to 37,715. This token reduction directly translates to a 19.3\% decrease in the reference self-attention computational load, dropping from 45.30 TFLOPs to 36.56 TFLOPs. Ultimately, these massive savings confirm that our mechanism critically streamlines the video-referencing pipeline, ensuring the architecture remains highly scalable and efficient for extended reference sequences.

\begin{table}[!t]
\caption{\textbf{Ablation on video reference audio.}}
\label{tab:ablation_audio}
\small
\centering
\setlength\tabcolsep{4pt}
\resizebox{0.60\textwidth}{!}{
\begin{tabular}{lcccc}
\hline
Variant & ID\textsubscript{ref}$\uparrow$ & ID\textsubscript{target}$\uparrow$ & Sync-C$\uparrow$ & Sync-D$\downarrow$ \\
\midrule
w/o Ref.\ Audio & 0.77 & 0.64 & 7.55 & 7.43 \\
Ours & \textbf{0.78} & \textbf{0.66} & \textbf{7.60} & \textbf{7.41} \\
\hline
\end{tabular}
}
\end{table}

\begin{table}[!t]
\caption{\textbf{Ablation on token selection efficiency.}}
\label{tab:ablation_token}
\small
\centering
\setlength\tabcolsep{4pt}
\resizebox{0.60\textwidth}{!}{
\begin{tabular}{lcc}
\hline
Variant & Avg.\ Tokens$\downarrow$ & FLOPs (RefSelfAttn)$\downarrow$ \\
\midrule
w/o Token Selection & 45,360 & 45.30 TFLOPs \\
Ours & \textbf{37,715} & \textbf{36.56 TFLOPs} \\
\hline
\end{tabular}
}
\end{table}

\begin{table}[h!]
\caption{\textbf{Ablation on masked DPO setting.}}
\small
\label{tab:ablation_dpo}
\centering
\setlength\tabcolsep{4pt}
\resizebox{0.7\textwidth}{!}{
\begin{tabular}{lccccc}
\hline
\multirow{2}{*}{Variant} & \multicolumn{2}{c}{Identity} & \multicolumn{2}{c}{Lip Sync} & \multirow{2}{*}{Overall$\uparrow$} \\
\cmidrule(lr){2-3} \cmidrule(lr){4-5}
& ID\textsubscript{ref}$\uparrow$ & ID\textsubscript{target}$\uparrow$ & Sync-C$\uparrow$ & Sync-D$\downarrow$ & \\
\midrule
DPO & 0.76 & 0.64 & 7.59 & 7.43 & 16.18 \\
Masked Real DPO & 0.73 & 0.61 & 6.60 & 8.10 & 13.07 \\
Masked DPO (Ours) & \textbf{0.78} & \textbf{0.66} & \textbf{7.60} & \textbf{7.41} & \textbf{16.42} \\
\hline
\end{tabular}
}
\vspace{-10pt}
\end{table}

\subsection{Ablation on DPO Setting}
\label{sec:abl_rl}
In our three-stage training strategy, we incorporate a task-specific reinforcement learning scheme implemented via Direct Preference Optimization (DPO)~\citep{rafailov2023direct}. Since the success of DPO heavily relies on how preference pairs (winning and losing samples) are constructed and optimized, we conduct an ablation study to validate our specific design choices. We compare our Masked DPO against two other variants: the standard unmasked DPO and a Masked Real DPO variant that strictly designates the ground-truth real video as the winning sample. Quantitative results are detailed in Table~\ref{tab:ablation_dpo}.

\noindent\textbf{Masked DPO.} We restrict the DPO loss exclusively to the human foreground to ensure the optimization process focuses entirely on the subject. As shown in Table~\ref{tab:ablation_dpo}, our `Masked DPO' outperforms the standard unmasked `DPO' variant, increasing ID\textsubscript{ref} from 0.76 to 0.78 and improving the ID\textsubscript{target} from 0.64 to 0.66. Without this spatial mask, the preference optimization is distracted by background pixels, weakening the network's capacity to refine critical facial identity features.

\noindent\textbf{Masked Real DPO.} We further investigate the source of the winning sample by comparing our approach, which constructs preference pairs from two model-generated outputs, against a Masked Real DPO baseline that utilizes the ground-truth video as the winning sample. Since our reinforcement learning objective is explicitly designed to maximize identity preservation, one might assume the real video provides the optimal target. However, Table~\ref{tab:ablation_dpo} reveals a paradox: using the real video actively degrades the identity fidelity, dropping ID\textsubscript{ref} to 0.73 and ID\textsubscript{target} to 0.61. This occurs possibly because the inherent distribution gap between raw real videos and synthesized outputs is too vast. Forcing the model to optimize against an unreachable, pixel-perfect ground truth destabilizes the preference learning process, confusing the identity reward. By constructing preference pairs exclusively within the model's generated distribution, our method ensures a stable optimization trajectory that successfully maximizes the targeted identity scores.

\subsection{Limitations}
\label{sec:limit}
There are several known limitations to our method. First of all, while TAVR excels at preserving facial identity and precise lip synchronization, the generation quality of highly articulate extremities, such as hands, can sometimes be suboptimal. Because our training objectives and token selection mechanisms heavily prioritize facial and head structural preservation, complex hand gestures may occasionally exhibit structural artifacts. Additionally, since our framework builds upon the Wan 2.1 foundation, it naturally inherits its specific architectural characteristics. Notably, the lower spatial-temporal compression rate of its underlying VAE requires processing larger latent representations, which inevitably leads to increased computational costs and higher GPU memory demands during both training and inference for extended sequences. Finally, our method primarily relies on aggregating visual cues from the provided reference context; therefore, generating the avatar under extreme target body poses or severe occlusions that are completely unobserved in the reference video may occasionally result in slight temporal inconsistencies or appearance degradation.

\end{document}